\documentclass[letterpaper,10pt,conference,romanappendices]{IEEEtran}
\IEEEoverridecommandlockouts  %

\usepackage{soul}
\usepackage{cite}
\usepackage{calc}
\usepackage{amsmath,amssymb,theorem}
\usepackage[export]{adjustbox}  %
\usepackage{graphicx}
\usepackage{mathtools}
\usepackage{cuted}
\usepackage{bbm}
\usepackage[ruled,vlined]{algorithm2e}
\usepackage{multirow}
\usepackage{epstopdf}	
\usepackage{xcolor}
\usepackage{color,xspace}
\usepackage{subfigure}
\usepackage{lipsum}
\usepackage{algpseudocode}
\usepackage{etoolbox}
\usepackage{inconsolata}
\usepackage{array}

\renewcommand{\baselinestretch}{0.9997}

\newcommand{\until}[1]{\{1,\dots, #1\}}

\newcommand{\timestep}{\Delta t}

\newcommand{\R}{\mathbb{R}}
\newcommand{\real}{\mathbb{R}}

\newcommand{\TwoNorm}[1]{\|#1\|}

\newcommand{\oprocendsymbol}{\hbox{$\bullet$}}
\newcommand{\oprocend}{\relax\ifmmode\else\unskip\hfill\fi\oprocendsymbol}

\newtheorem{problem}{Problem}

\newcommand{\vmax}{v_\text{max}}
\newcommand{\omegamax}{\omega_\text{max}}
\newcommand\smdots{\ifmmode\ldots\else\makebox[0.5em][c]{.\hfil.\hfil.}\thinspace\fi} %

\newcolumntype{L}[1]{>{\raggedright\let\newline\\\arraybackslash\hspace{0pt}}m{#1}}
\newcolumntype{C}[1]{>{\centering\let\newline\\\arraybackslash\hspace{0pt}}m{#1}}
\newcolumntype{R}[1]{>{\raggedleft\let\newline\\\arraybackslash\hspace{0pt}}m{#1}}

\begin{document}

\makeatletter
\newcommand{\linebreakand}{%
  \end{@IEEEauthorhalign}
  \hfill\mbox{}\par
  \mbox{}\hfill\begin{@IEEEauthorhalign}
}
\makeatother

\author{\IEEEauthorblockN{
Kevin Zhu
}
\IEEEauthorblockA{
\textit{George Mason University}\\
Fairfax, USA \\
kzhu4@gmu.edu}

\and
\IEEEauthorblockN{
Connor Mattson
}
\IEEEauthorblockA{
\textit{University of Utah}\\
Salt Lake City, USA \\
c.mattson@utah.edu}

\and
\IEEEauthorblockN{
Shay Snyder
}
\IEEEauthorblockA{
\textit{George Mason University}\\
Fairfax, USA \\
ssnyde9@gmu.edu}

\and
\IEEEauthorblockN{
Ricardo Vega
}
\IEEEauthorblockA{
\textit{George Mason University}\\
Fairfax, USA \\
rvega7@gmu.edu}

\linebreakand
\IEEEauthorblockN{
Daniel S. Brown
}
\IEEEauthorblockA{
\textit{University of Utah}\\
Salt Lake City, USA \\
daniel.s.brown@utah.edu}

\and
\IEEEauthorblockN{
Maryam Parsa
}
\IEEEauthorblockA{
\textit{George Mason University}\\
Fairfax, USA \\
mparsa@gmu.edu}

\and
\IEEEauthorblockN{
Cameron Nowzari
}
\IEEEauthorblockA{
\textit{George Mason University}\\
Fairfax, USA \\
cnowzari@gmu.edu}
}

\title{Spiking Neural Networks as a Controller for Emergent Swarm Agents

}
      
\maketitle

\begin{abstract}
Drones which can swarm and loiter in a certain area cost hundreds of dollars, but mosquitos can do the same and are essentially worthless.
To control swarms of low-cost robots, researchers may end up spending countless hours brainstorming robot configurations and policies to ``organically" create behaviors which do not need expensive sensors and perception.
Existing research explores the possible emergent behaviors in swarms of robots with only a binary sensor and a simple but hand-picked controller structure. Even agents in this highly limited sensing, actuation, and computational capability class can exhibit relatively complex global behaviors such as aggregation, milling, and dispersal, but finding the local interaction rules that enable more collective behaviors remains a significant challenge.
This paper investigates the feasibility of training spiking neural networks to find those local interaction rules that result in particular emergent behaviors. 
In this paper, we focus on simulating a specific milling behavior already known to be producible using very simple binary sensing and acting agents. 
To do this, we use evolutionary algorithms to evolve not only the parameters (the weights, biases, and delays) of a spiking neural network, but also its structure.
To create a baseline, we also show an evolutionary search strategy over the parameters for the incumbent hand-picked binary controller structure. Our simulations show that spiking neural networks can be evolved in binary sensing agents to form a mill.
\end{abstract}


\section{Introduction}

Ants appear to be fairly simple creatures on the surface, but they are capable of many group behaviors, such as bridging, or perhaps more interestingly, milling, in which groups of ants march in a circle until they die. Ant mills form circumstantially, in large part due to the pheromones left behind by other ants \cite{franksBlindLeadingBlind1991a}.
These behaviors are emergent, that is, the behavior is a property of the entire swarm, resulting from local interaction rules that do not explicitly define the global swarm state. 
Ant milling is an emergent behavior; a tragic consequence of each ant's internal interaction policy rather than a behavior learned through incentives.

To replicate this behavior in a robot swarm, no explicit communication or pheromones are required. We restrict our interest to an even more limited sensor suite: a single binary sensor which activates if another object passes within its field of view. Deceptively simple controllers on binary sensing-to-actuating robots have been shown to be capable of what were previously thought to be complex global behaviors requiring good sensing and localization capabilities. Take, for instance, a simple binary controller that maps the binary sensor output to one of two constant motor commands. With certain sets of motor commands, swarms of simple ground robots are capable of aggregation, dispersion, and milling 
\cite{gauciEvolvingAggregationBehaviors2014, brownDiscoveryExplorationNovel2018}.

This simple binary controller was originally designed by Gauci et al., and evolutionary algorithms were used to discover parameters for the controller that lead to milling in \cite{gauciEvolvingAggregationBehaviors2014}. Despite its simplicity, this controller has been shown to be robust to sensing and actuation noise \cite{gauciEvolvingAggregationBehaviors2014, daymudeDeadlockNoiseSelfOrganized2021}.

We propose searching for controller structure and parameters simultaneously. The architecture of the controller should lend itself to the learning of essential skills in robotics, such as perception, prediction, planning, and action. We look at spiking neural networks as a controller for emergent behaviors in multi-agent systems.
In this paper, we examine searching for controller structures that produce a milling behavior, which provides a baseline for future work on more complex behaviors.

Our contributions are as follows:
\begin{itemize}
    \item We present a framework for training a swarm to exhibit a specific emergent behavior, milling, with minimal intuition and without explicitly defining the controller structure.
    \item We show that an evolutionary algorithm can be used to train a spiking neural network to produce milling in an agent-based swarm simulator.
\end{itemize}

\section{Background}

\subsection{Spiking Neural Networks for Multi-Agent Systems}

The third generation of neural networks take further inspiration from the human brain \cite{maassNetworksSpikingNeurons1997} to create a more energy efficient and noise resilient deep learning system operating on spikes~\cite{schuman2022opportunities}. More formally known as spiking neural networks (SNNs), these systems differ from traditional ANNs by propagating spikes through the network rather than floating-point values. In particular, the Leaky Integrate-and-Fire model of neuron incorporates neuronal charge, passive leaking of said charge over time, and variable delay of spike transmission between neurons~\cite{DBLP:journals/corr/SchumanPPBDRP17}. Perhaps the most popular motivation for studying SNNs is their potential use in size, weight, and power constrained applications. There is ongoing hardware development to make this a reality~\cite{DBLP:journals/corr/SchumanPPBDRP17}, but such systems have not been widely deployed yet. SNNs are also theorized to be more tolerant to faults and noise owing to their asynchronicity and parallel processing capabilities~\cite{DBLP:journals/corr/SchumanPPBDRP17}. Furthermore, this neuromorphic computing framework has lead to entire new areas of computer and neural architecture design for various forms of computation~\cite{aimone2022review} and optimization~\cite{10.1145/3589737.3605998}.

\textit{Our} primary interest in SNNs lies in their inbuilt notion of time and their intermingled memory owing to the charge accumulated on each neuron. Some variants of ANNs, such as the Long-Short Term Memory (LSTM) architecture, do have memory capabilities, and have successfully been applied to various problems in robotics, such as path planning~\cite{nicolaLSTMNeuralNetwork2018}. However, the structure of an LSTM cell is complex and was painstakingly designed by researchers, and contains no inbuilt notion of time. For example, scheduling an event to occur in the future would require an LSTM to be given an explicit timestamp upon each cycle, or require an accumulator-like structure to be learned within the network. Compare this to SNNs, where a spike can simply be delayed. We believe that these properties of SNNs lend themselves to memory, prediction, and pre-acting, which we see as essential components of complex behaviors.

In the scope of robotics, other works show SNN models matching or out-performing ANNs in single and multi-agent scenarios~\cite{saravananExploringSpikingNeural2021b} and have also been deployed as a collision avoidance decision mechanism in~\cite{zhao2022nature}.
SNNs are an appealing candidate for autonomous robotics, but training them requires novel methods because their activation function is nonlinear and time-dependent.
Training approaches include translating pre-trained ANNs to SNNs~\cite{caoSpikingDeepConvolutional2015}, using differentiable surrogates for gradient descent learning ~\cite{leeEnablingSpikeBasedBackpropagation2020,
shresthaSLAYERSpikeLayer2018a,
neftciSurrogateGradientLearning2019a} and evolutionary algorithms \cite{dasTrainingSpikingNeural2021, schumanEvolutionaryVsImitation2022} paired with Bayesian optimization \cite{parsa2020bayesian, parsa2019bayesian}.
Evolutionary algorithms have also been used to design the architecture and train the weights of an SNN to process LIDAR data and control a 4-wheeled robotics platform \cite{patton2021neuromorphic}.
In \cite{dasTrainingSpikingNeural2021}, a novel genetic algorithm was used to evolve a multi-agent system to compete for food. 

\subsection{Evolutionary Optimization for Neuromorphic Systems}
Evolutionary algorithms (EAs) are good at \textit{exploring} the search space, but tend to be inefficient for \textit{exploiting} and fine-tuning.
Despite their inefficiencies, there are several reasons why EAs are suited to our problem.

\textbf{1. }Compared to gradient based methods, EAs are extremely flexible as the reward structure can be sparse and non-differentiable.

\textbf{2. }EAs have been used to evolve neural network structure alongside the typical weights and biases~\cite{stanley2002evolving}.

In this paper, we use TENNlab's \cite{plankTENNLabExploratoryNeuromorphic2018} EONS evolutionary framework \cite{schumanEvolutionaryOptimizationNeuromorphic2020a} and virtual Caspian SNN processors \cite{mitchellCaspianNeuromorphicDevelopment2020} to exploit Darwin's Theory of Evolution \cite{darwin1859} and design a controller that can be homogenously applied to a swarm of binary sensing agents to achieve a milling behavior. Like \cite{stanley2002evolving}, EONS is capable of evolving not only the weights and hyperparameters of a network, but also its structure. This technique results in model configurations that resemble heterogeneous graph structures~\cite{zhou2020graph} along with having a multitude of unique performance attributes compared to traditional layer-based systems~\cite{hosseini2020deep}.

To the best of our knowledge, this is the first work applying SNNs to a binary sensing swarm.  %

\section{Problem Formulation}\label{sn:problem_formulation}

We consider a swarm of~$N$ unicycle-like robots with discretized kinematics for the $i$-th robot with~$i \in \until{N}$
\begin{align}
x_i(k+1) &= x_i(k) + u_{i,1} \cos \theta_i \timestep \\
y_i(k+1) &= y_i(k) + u_{i,1} \sin \theta_i \timestep \\
\theta_i(k+1) &= \theta_i(k) + u_{i,2} \timestep ,
\end{align}

where~$z_i(k) = (x_i,y_i,\theta_i) \in \operatorname{SE}(2)$ 
represent the 2D position and orientation of agent~$i$. Each robot has two control inputs in the form of a forward velocity~$u_{i,1} = v_i \in [-\vmax, \vmax]$ and a turning rate~$u_{i,2} = \omega_i \in [-\omegamax,\omegamax].$

We consider a binary sensing model where agent~$i$ receives the observation~$h_i \in \{0,1\}$ given by
\begin{align}\label{eq:output}
    h_i(z) = \begin{cases}
                 1 & \text{if } \exists j \neq i, s.t. ~z_j \in \operatorname{FOV}_i , \\
                 0 & \text{otherwise,}
             \end{cases}
\end{align}
where the $\text{FOV}_i$ is the conical area in front of the sensor of the $i$th agent with a measured distance and opening angle, as shown in Fig.~\ref{fig:FOV_explainer}.

\begin{figure}
    \centering
    \includegraphics[max height=4cm]{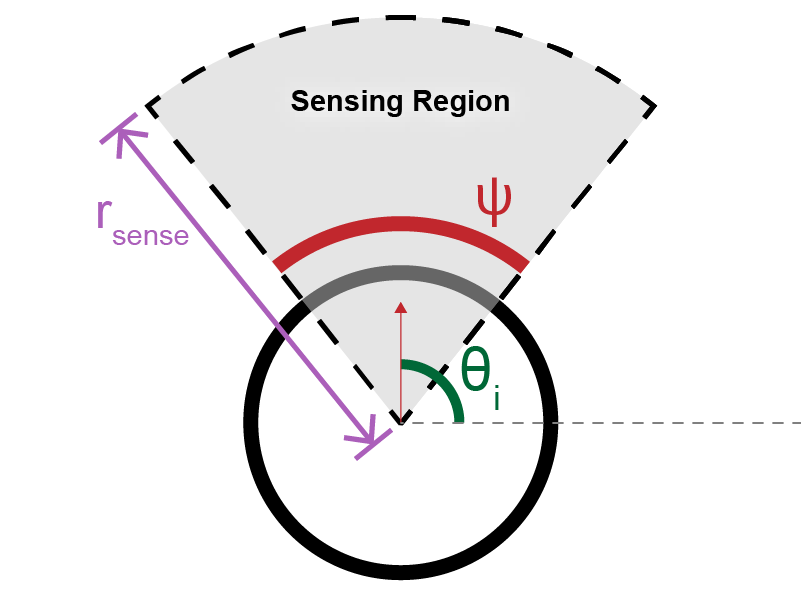}
    \caption{Diagram explaining sensor variables and sensor placement}
    \label{fig:FOV_explainer}
\end{figure}

Since the goal of this paper is not to find arbitrary emergent behaviors, but train networks to optimize a particular emergent behavior, we must choose a fitness function. 

In \cite{ramosEvolvingFlockingEmbodied2019}, it is suggested that rather than calculating rewards from an agent-local perspective, using a globally defined metric to evolve the local behaviors can lead to successful flocking behaviors. To that end, we use the ``circliness" value, first described in \cite{taylorImpactCatastrophicCollisions2021}. This is a measure of how perfect of a circle is formed by the agents, which we take to be the fitness of a milling swarm.

We begin with a ``fatness" measure $\phi$ defined in \cite{taylorImpactCatastrophicCollisions2021}. Note that the following equations \eqref{eq:fatness}, \eqref{eq:tangentness} are defined for a particular state and time $z(k)$, which we omit for readability. Letting $\mu$ be the average position of all agents and $\mathbf{p}_i = (x_i, y_i) \in \R^2$ be the 2D position of the $i$-th agent,
$$\mu = \frac{1}{N}\sum^{N}_{i=1}\mathbf{p}_i$$
and $r_\text{min}, r_\text{max}$ be the minimum and maximum distance to the center of the swarm $\mu$,
\begin{align*}
    r_\text{min} &= \min_{i \in \{1, \dots, N\}} \| \mathbf{p}_i - \mu \| , \quad
    r_\text{max} &= \max_{i \in \{1, \dots, N\}} \| \mathbf{p}_i - \mu \|
\end{align*}
\begin{align}\label{eq:fatness}
\phi = 1 - \frac{r^2_\text{min}}{r^2_\text{max}}
\end{align}

As noted in \cite{taylorImpactCatastrophicCollisions2021}, $\phi$ is a measure of radial variance. $\phi = 1$ indicates that the agents lie perfectly on a circle at time $k$, and $\phi \rightarrow 0$ implies radial scatter.

Instantaneous tangentness is taken as:
\begin{align}\label{eq:tangentness}
    \tau &= \frac{1}{N} \sum_{i=1}^N \vert \sigma_\tau  \vert,  \\
    \text{where } \sigma_\tau &= \frac{\mathbf{p}_i - \mu}{\TwoNorm{\mathbf{p}_i - \mu}} \cdot \frac{\dot{\mathbf{p}}_i}{\TwoNorm{\dot{\mathbf{p}}_i}} \label{eq:tangentnes_vec}\\
     &= \cos (\theta_i - \angle (\mathbf{p}_i - \mu))
     \label{eq:tangentness_cos}
\end{align}
Thus, $\tau=0$ indicates that agent headings are perfectly perpendicular to the centroid, implying tagentness to the circle formed by the agents, and $\tau = 1$ represents all agents are facing into or away from the centroid $\mu$.
Note that instead of \eqref{eq:tangentnes_vec} from \cite{taylorImpactCatastrophicCollisions2021}, we use \ref{eq:tangentness_cos} to avoid division by zero; equivalent in all other cases.

It is desired that the agents maintain this formation for as much of the time as possible, as opposed to a perfect circle forming for a brief instant and then dissipating. To encourage this, we use the averaged fatness and tangentness --- $\; \bar{\phi} (k), \, \bar{\tau} (k)$ --- over the previous $T_\text{window}$ ticks i.e. over the rolling interval $[k - T_\text{window}, \, k]$ where $k \geq T_\text{window}$.
We can then use the circliness metric defined in \cite{taylorImpactCatastrophicCollisions2021}:
\begin{align}\label{eq:circliness}
\lambda(k) = 1 - \max \{ \bar{\phi}(k), \bar{\tau}(k) \} 
\end{align} where $\lambda \in [0, 1]$. $\lambda = 1$ corresponds to a swarm of agents travelling on a single circular path, and $\lambda = 0$ ``represents maximum disorder" as noted in \cite{taylorImpactCatastrophicCollisions2021}.

Our goal now is to find controllers~$u_{i,1}$ and~$u_{i,2}$ to make circliness~$\lambda(T)$ as close to $1$ as possible. We formalize this next.

Although we consider an unbounded domain~$\mathbf{p}_i \in \real^2$, due to practical limitations such as limited sensing range and collisions, it is sensible to restrict the set of initial conditions. 
In particular, we initialize all agent positions~$\mathbf{p}(0)$ randomly in a square starting area with random heading, $\theta_i(0) \sim \mathcal{U}[0, 2\pi)$.
\begin{align}\label{eq:starting_region}
x_i(0), y_i(0) \sim \operatorname{U}[-W/2, W/2]  \text{ where } W \in \real
\end{align}
where W defines the width and height of the starting region.
Additionally, no agents are overlapping each other i.e. $ z_0 \in \{z_0 \, \vert \; \| \mathbf{p}_i - \mathbf{p}_j \| > 2 r \}$ for $i, j \in \{1, \dots, N\} , j \neq i$

\begin{problem}[Optimizing Milling]\label{pr:main}
{\rm
Given an initial condition~$z_0 \in \operatorname{S}$, find controllers~$u_{i,1},u_{i,2}$ as functions of the binary observation of each agent~$h_i$ to solve
\begin{align*}
\begin{array}{cl} 
\underset{ \pi_1, \pi_2 }{\text{maximize}} & \lambda(T), \\ \text{subject to} & z(T) = z(0) + \timestep \sum_{k=0}^{T-1} f(z(k),u(k))  , \\ 
& z(0) = z_0, \\
& u_{i,1}(k) \in [-\vmax,\vmax] \: \forall \, k \in  \until{T-1} , \\
& u_{i,2}(k) \in [-\omegamax,\omegamax] \: \forall \, k \in \until{T-1},\end{array}
\end{align*}
where~$\pi_1 : \{0,1\}^* \rightarrow u_{i,1}$ and~$\pi_2 : \{0,1\}^* \rightarrow u_{i,2}$ are control policies driving the forward speeds~$v_i$ and turning rates~$\omega_i$ of each agent, respectively,
where $*$ is the Kleene Star, denoting ``zero or more repetitions" of items from the set \cite{hopcroftRegularExpressions1979}. 
Thus, the policies $\pi_1$ and $\pi_2$ take as input
the history of all binary network inputs since the robot was turned on. As in \cite{vega2023simulate, snyder2023zespol}, we assume that the controllers $\pi_1, \pi_2$ are homogeneous across the swarm.
}

\end{problem}

\begin{center}
\end{center}

\section{Methods}

To evaluate the capabilities of an SNN in creating an emergent milling behavior, we compare this to a symbolic controller described in \cite{gauciEvolvingAggregationBehaviors2014}: a direct binary sensing-to-action controller, which can be formally defined as
\begin{align}\label{eq:binary_s2a_controller}
u_{i,2}(h_i), \: u_{i,1}(h_i) &= \begin{cases}
    v_a, \: \omega_a \quad & \text{if } h_i = 1 \\
    v_b, \: \omega_b \quad & \text{otherwise.} \end{cases}
\end{align}
While this controller can be stated in a pure fashion i.e. $u_{i, 1}(k)$ depends only on $h_i(k)$, it is more conducive to simulate it as $u_{i, 1}(k) = f_1(h_i(k-1))$. While the controller structure is clearly defined in \eqref{eq:binary_s2a_controller}, the parameters $v_a, v_b, \omega_a, \omega_b$ must be chosen to determine the behavior of the swarm. It has been shown to result in six unique behaviors including milling, referred to as ``circular pursuit" in \cite{brownDiscoveryExplorationNovel2018}. To determine a set of parameters for our baseline, we use genetic algorithms, particularly CMA-ES, further discussed in Section \ref{sn:cmaes}. 

A controller using an SNN must be described differently:
\begin{align}\label{eq:snn_controller}
u_{i,1}(k), \, u_{i,2}(k), \, S_i(k) &= f_\text{SNN}(h_i(k-1),\, S_i(k-1))
\end{align}
where $s_i(k)$ is the state of the network of agent $i$ at time $k$. Note that the controller can no longer be stated as a pure function; its output depends on the previous state of the SNN, owing to the intermingled computing and memory.

The structure and parameters of $f_\text{SNN}$ must be defined. Furthermore, as the raw inputs and outputs of the spiking neural network are spikes rather than numerical values, they must be converted. We detail our methods for both of these problems in this section.

\subsection{Encoder and Decoder}
The binary output of the sensor must be encoded to a spike train to be sent to the SNN. To do this, we first convert the binary output of $h_i(k)$ using one-hot encoding (i.e. $0 \rightarrow [1, 0]^T$, $1 \rightarrow [0, 1]^T$). This is a type of ``binning" as described in \cite{schumanNonTraditionalInputEncoding2019}. Each element of the vector is then converted to a spike train to be given to its respective input neuron: If the neuron is to receive a $1$, a spike is fed to that input neuron on every cycle of the simulated Caspian processor containing the SNN; See Figure \ref{fig:encoders} and section \ref{sn:caspian_lif}. The processor is then run for 3 cycles to account for spike propagation time, and then run for another 10 cycles.

\begin{figure}%
    \centering
    \includegraphics[width=1\linewidth]{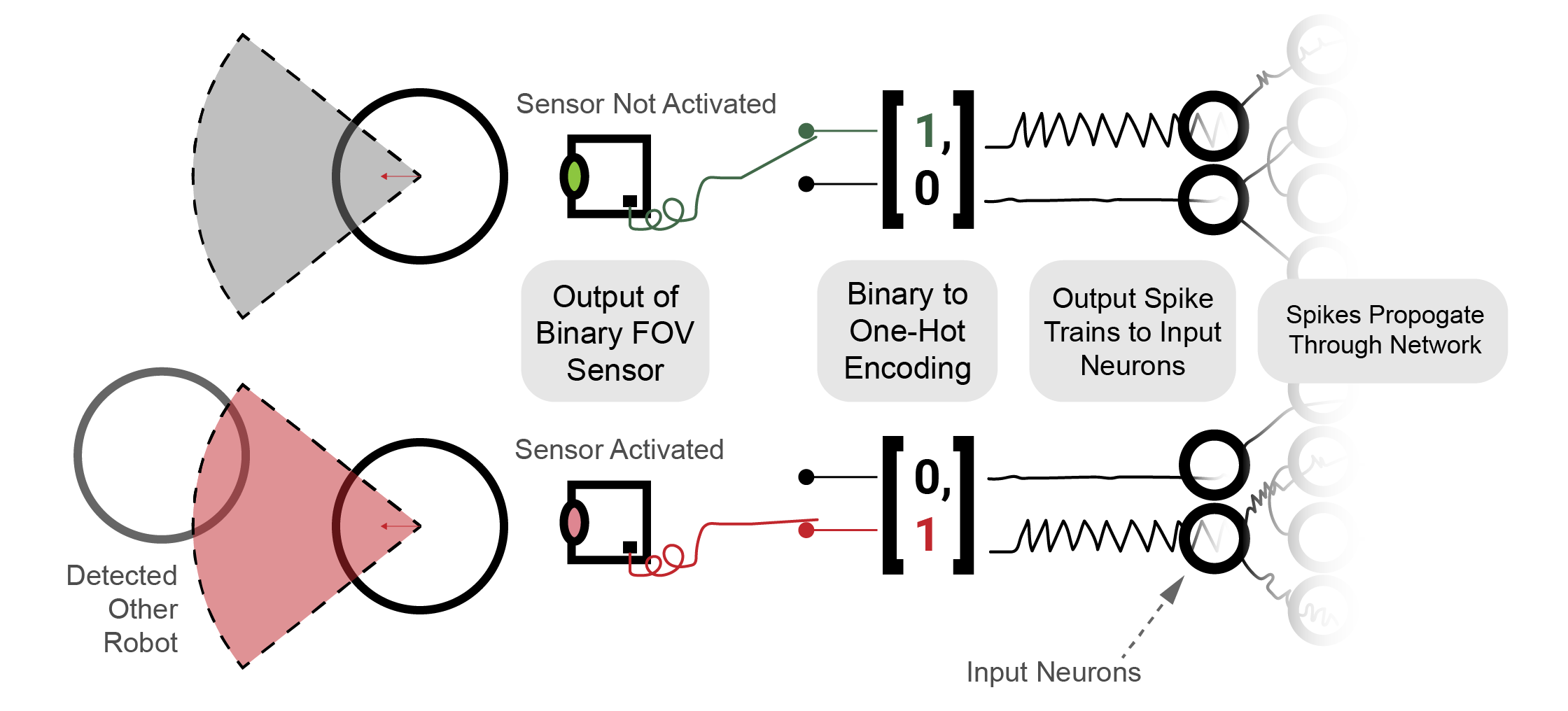}
    \caption{Diagram of encoding process for a single robot. Top: Sensor not activated; robot does not ``see" another agent. Bottom: Sensor activated, robot ``sees" other agent in its FOV.}
    \label{fig:encoders}
\end{figure}

During these cycles, the spikes are propagated through the network, and some spikes will likely be emitted from the output neurons. The arrival time and magnitude of these spikes are recorded by the decoder. For our decoder, we use a rate-based encoding. The total number of spikes emitted from an output neuron during the last 10 cycles is summed. This value is then normalized by dividing by 10. As such, each output neuron can output 10 distinct discrete values. We use 4 output neurons; To represent negative values, we used the difference in pairs of outputs, i.e. $v = o_1 - o_2$, which allows for a maximum of 21 distinct values per pair. As this value is normalized,
\begin{equation}\label{eq:decoder_range}
    o \in \{-1, \smdots, 0, 0.1, \smdots, 1 \}
\end{equation}
we multiply this by the maximum safe forward velocity $v_\text{max}$ of the robot to obtain $u_{i,1}(k+1)$. Another pair of output neurons is used for the turning rate $\omega = o_3 - o_4$, scaled to the robot's maximum angular velocity $\omega_\text{max}$, thus we have $u_{i,2}(k+1)$.

\subsection{Evolving the Network} \label{me:eons}  %
As discussed in the previous section, the number of input and output neurons are fixed at two and four respectively. However, these neurons and indeed the remaining internal neurons have parameters that must be determined. Furthermore, we do not have a fully-connected internal network structure, thus the network edges and edge parameters must also be determined. To do this, we use TENNlab's \cite{plankTENNLabExploratoryNeuromorphic2018} EONS \cite{schumanEvolutionaryOptimizationNeuromorphic2020a} library which implements genetic algorithms capable of creating and evolving the network structure and parameters for our SNN.

We now describe the search process. First, $P$ networks are initialized. The two input and four output neurons are created first. Each network is given $n_\text{hidden}$ ``floating" unconnected nodes. Then, $n_\text{edges}$ edges are randomly assigned. Each node and edge's parameters are also randomized within limits set for each parameter (see Caspian parameters in Table \ref{ta:caspian_params}). Each initial network is likely to be unique to the others in the population.

We use tournament selection~\cite{schumanEvolutionaryOptimizationNeuromorphic2020a}, with the $n_\text{best}$ best networks being directly copied to the next generation. The parameters of random edges and nodes are mutated, and nodes and edges may be added or removed, as described in \cite{schumanEvolutionaryOptimizationNeuromorphic2020a}. A certain percentage of the population, determined by \texttt{random\_factor}, is replaced with newly initialized networks.

\subsection{Simulation}

To calculate the fitness of a particular network, a simulator instance is instantiated. The starting positions and headings $z_0$ of the $N$ agents are set from the uniform distribution discussed in the problem formulation, but with a fixed seed, resulting in consistent initial state across all training runs. This is done to prevent variations in the circliness $\lambda$ due to differing initial conditions. An increase in the circliness score then is purely due to an improvement in behavior, at least for that particular seed. $N$ simulated Caspian processors are created and loaded with a copy of the network structure. The simulator and simulated Caspian SNN processors are then run in the observe-act loop for $T$ timesteps, and the static metrics fatness $\phi$ and tangentness $\tau$ are calculated and stored. Finally, the circliness value $\lambda(T)$ from eq. \eqref{eq:circliness} is used as the reward for that particular run and its associated network.

We performed this process in parallel on a multi-core workstation, and each thread is assigned the fitness calculation of a particular network. After each generation, if the fitness of the best network is greater than that of the previous best, the network is saved. Thus the output of the search is the first network to reach the maximum fitness score, if the maximum score is achieved.

\subsection{Caspian} \label{sn:caspian_lif}

To process the SNN, we use a simulated TENNlab Caspian processor \cite{mitchellCaspianNeuromorphicDevelopment2020}, which uses a leaky-integrate-and-fire (LIF) neuron model. Each neuron can take four parameters: leak, axonal delay, threshold, and weight. Each synapse (edge) has two parameters: weight and synaptic delay. When a neuron receives a spike, the magnitude of the spike is added to that neuron's charge. Over time as specified by the leak parameter, a neuron's charge will decay to zero. If the charge reaches the threshold, the neuron will spike and the charge will be reset to resting state. A neuron can fire once per processor tick. 

\section{Experimental Setup}

We used the parameters in Tables \ref{ta:common_params}, \ref{ta:eons_params}, \ref{ta:caspian_params} for every experiment, except where otherwise noted.
\renewcommand{\arraystretch}{1.1}
\begin{table}
\begin{center}
\caption{Base Parameters for all experiments (unless otherwise noted)}\label{ta:common_params}
\begin{tabular}{|lrll|}
\hline
Variable & Value & Unit & Description \\
\hline
\multicolumn{3}{|l}{Problem Parameters}&\\
$N$ & $ 10 $ & & Number of Agents \\
$W$ & $1.2$ & meters & Spawn Area Width \\  %
$T$ & $1000$ & ticks & Time horizon \\ 
$T_\text{window}$ & $450$ & ticks & Rolling Window Horizon \\
$\timestep$ & $ 1/7.5$ & s & timestep \\
$\vmax$ & $ 0.2 $ & m/s & Max Forward Velocity \\
$\omegamax$ & $ 2.0 $ & rad/s & Max Angular Velocity \\
$r_\text{sense}$ & $3.6$ & meters & Sensing Range\\
$\psi$ & $0.4$ & rad & Sensor FOV \\
\hline
\multicolumn{3}{|l}{Training Parameters}&\\
$P$ & 100 & & Population Size \\
$n_\text{epochs}$ & $1000$ & & Max \# of Epochs \\
\hline
\end{tabular}
\end{center}
\end{table}

\subsection{RobotSwarmSimulator (RSSim)}
We used a Python-based 2D agent simulator to test both the baseline controller and our SNN approach: RobotSwarmSimulator (RSSim)~\cite{mattson2023leveraging}.
RSSim uses the agent and sensor dynamics detailed in Section \ref{sn:problem_formulation} to update the position of each simulated robot each tick based on the control inputs. Collisions are resolved inelastically. In the case of the SNN, on each tick of RSSim, each robot evaluates its sensor state and sends the binary value through the encoder. An emulated Caspian processor executes the SNN, and its output is decoded, from which the agent's desired position is calculated. In the case of the symbolic controller, the binary sensor output is used to choose between the two sets of actuation outputs as shown in \eqref{eq:binary_s2a_controller}, the values of which are determined using CMA-ES.

\subsection{CMA-ES} \label{sn:cmaes}
To evolve the symbolic controllers in RSSim, we use Covariance Matrix Adaptation - Evolutionary Strategy (CMA-ES) \cite{hansen2001completelyDerandomizedSelfAdaptation, hansen2016CMAEvolutionStrategy}, a form of evolutionary optimization that samples a population from a multivariate Gaussian distribution,
$X_i \sim \mathcal{N}_k(\epsilon,\,\sigma^{2}C_k)$,
and updates the mean~$\epsilon$ and covariance matrix each epoch as a way to optimize a black-box function $f(x_1, x_2, ..., x_k)$ with $k$ parameters.
We use CMA-ES to search for a pair of binary sensing-to-action controllers  \eqref{eq:binary_s2a_controller} that maximizes the fitness function  $\lambda(T)$. Specifically, $v_a, v_b, \omega_a, \omega_b$ are used as the search parameters ($k$=4), each normalized from the max forward and angular velocity limits to the range $[0, 1]$. We set the initial starting mean $\epsilon_0 = [0.5, 0.5, 0.5, 0.5]$, and step-size, $\sigma_0=0.2$. During each epoch, population candidates are simulated in parallel and the fitness is returned to CMA-ES, which updates the mean, covariance matrix, and step-size for the distribution of the subsequent epoch. For a fair comparison, CMA-ES uses the same population size, $P$, and iteration cutoff, $n_{\text{epochs}}$ as SNN evolutionary search. All other hyper-parameters follow the standard CMA-ES defaults \cite{nikolaus_hansen_2023_7573532}.

CMA-ES is configured to output continuous values for each search parameter within the range [0, 1]. To match the discrete nature of the SNN decoder output, \eqref{eq:decoder_range}, we round each output to the nearest 0.05, resulting in the discrete set $\{0.0, 0.05, \dots, 0.95, 1.0\}$ (21 values). Discretized candidate solutions are then mapped back to the appropriate control signal in the ranges $[-\vmax,\vmax]$, $[-\omegamax,\omegamax]$ to obtain $u_{i,1}$, $u_{i,2}$ respectively, as in \eqref{eq:binary_s2a_controller}.

\subsection{EONS}\label{sn:eons-methods}

For our SNN-based approach, EONS \cite{schumanEvolutionaryOptimizationNeuromorphic2020a} was used with both simulators to search for a controller that maximizes $\lambda(T)$. The EONS parameters used are shown in Table \ref{ta:eons_params} unless otherwise noted. EONS is restricted to generating networks with parameters limited to the parameter space described in Table \ref{ta:caspian_params}. Outside of this, EONS is not processor-aware; it has no knowledge of the parameters themselves.

{
\renewcommand{\arraystretch}{1}
\begin{table}
\begin{center}
\caption{EONS parameters used}\label{ta:eons_params}
\begin{tabular}{|lrl|}
\hline
Variable & Value & \\
\hline
\texttt{starting\_nodes} & 10 & \\
\texttt{starting\_edges} & 20 & \\
\texttt{population\_size} & $P$ & \\
\texttt{crossover\_rate} & 0.5 & \\
\texttt{mutation\_rate} & 0.9 & \\
\texttt{add\_node\_rate}$^*$ & 0.55 & \\
\texttt{delete\_node\_rate}$^*$ & 0.45 & \\
\texttt{add\_edge\_rate}$^*$ & 0.6 & \\
\texttt{delete\_edge\_rate}$^*$ & 0.4 & \\
\texttt{selection\_type} & ``tournament" & \\
\texttt{tournament\_size\_factor} & 0.1 & \\
\texttt{tournament\_best\_net\_factor} & 0.9 & \\
\texttt{random\_factor} & 0.10 & \\
\texttt{num\_mutations} & 3 & \\
\texttt{node\_mutations} & { ``Threshold": 1.0 } & \\
\multirow{2}{*}{\texttt{edge\_mutations}} & ``Weight": 0.65 &\\
& ``Delay": 0.35 & \\
\texttt{num\_best} & 4 & \\
\hline
\multicolumn{3}{p{6cm}}{$^*$ Add/delete parameters are relative to each other and sum to $1.0$.} \\
\end{tabular}
\end{center}
\end{table}}

\begin{table}%
\begin{center}\caption{Range of Caspian parameters that EONS was allowed to generate.}\label{ta:caspian_params}
\begin{tabular}{|lr|}
\hline

Parameter & Range$^{\mathrm{a}}$ \\
\hline
Threshold & $[0, 127]$ \\
Leak Time Constant & None, 1, 2, 4, 8, or 16  \\
Axonal Delay & 0 \\
Synaptic Weight & $[-127, 127]$ \\
Synaptic Delay & $[0, 255]$ \\
Cycles per Tick$^{\mathrm{b}}$ & $ 13 $ \\
\hline
\multicolumn{2}{l}{$^{\mathrm{a}}$ Only integer values are allowed.} \\
\multicolumn{2}{L{6cm}}{$^{\mathrm{b}}$ refers to the number of Caspian cycles corresponding to a single sense-act `tick' in RSSim.}
\end{tabular}
\end{center}

\end{table}

We first generate a population of $P$ networks. For each node and edge in the network, each parameter is initialized to a uniform random value within that parameter's minimum and maximum values, shown in Table \ref{ta:caspian_params}. Each network in this population is evaluated by running parallel simulations where that network is loaded into each robot in the swarm. The final circliness value, $\lambda(T)$, is used to compute the fitness for that network. Neurons and synapses not part of the input/output path were not explicitly pruned; rather, a penalty was subtracted from $\lambda(T)$ based on the number of neurons and synapses for that network, following the multi-objective optimization described by Schuman et al. \cite{schumanEvolutionaryOptimizationNeuromorphic2020a}. EONS then evolves the population for the next epoch.

\subsection{Training}

A single CMA-ES training run was conducted. For EONS, due to the variability of the training process, five runs were conducted across different seeds for EONS' random number generator. For both EONS and CMA-ES, following each epoch, the system time, epoch number, and the circliness values across the population are recorded. The search processes were performed on the same desktop computer with an AMD Ryzen 9 3900 desktop-class CPU running a Debian-based OS inside a ProxMox virtual machine. No other tasks were running during the search process. All evolutionary circliness evaluation was carried out with 24 threads (i.e. a max of 24 concurrent simulations).

\section{Results}
\subsection{Training Fitness}

\begin{figure}%
    \centering
    \includegraphics[width=.9\linewidth]{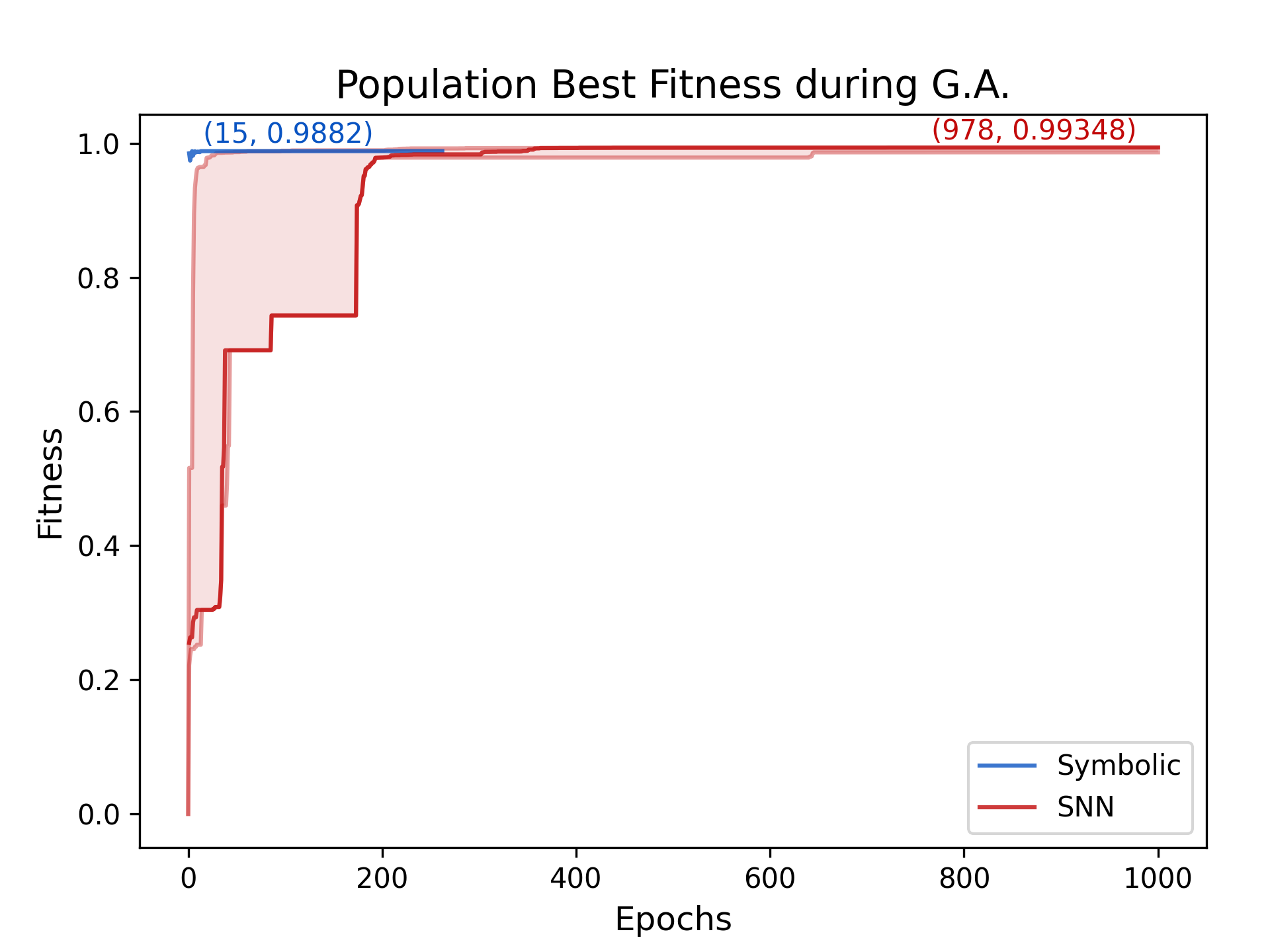}
    \includegraphics[width=.9\linewidth]{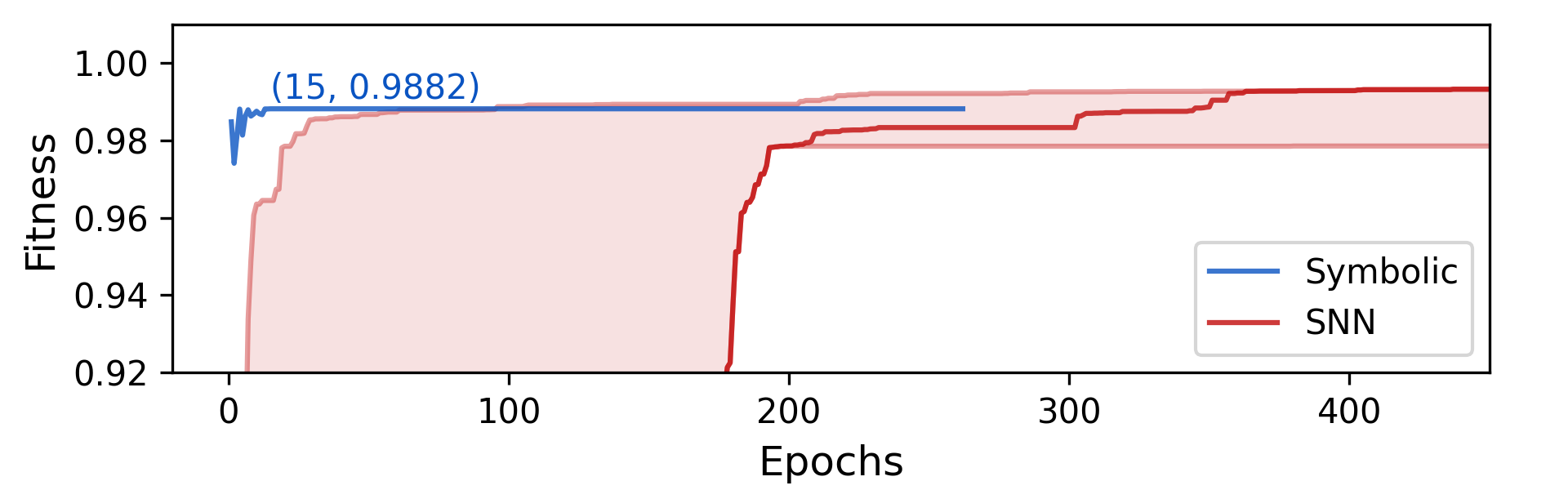}
    \caption{
    \textbf{Top:~}Population Best Fitness per Epoch. For the symbolic 4-parameter controller optimized with CMA-ES, a single run was performed, and the search was terminated at the 263rd epoch. For the SNNs evolved with EONS, 5 runs with different evolutionary seeds were performed. The shaded silhouette bounds the min/max fitness achieved for a particular epoch across the EONS runs. The EONS run which achieved the highest final fitness is highlighted. The annotations indicate when the highest fitness score was first achieved in the best run. \\
    \textbf{Bottom:~}Focused view of Top plot.
    }
    \label{fig:bests}
\end{figure}

\begin{figure}%
    \centering
    \includegraphics[width=.9\linewidth]{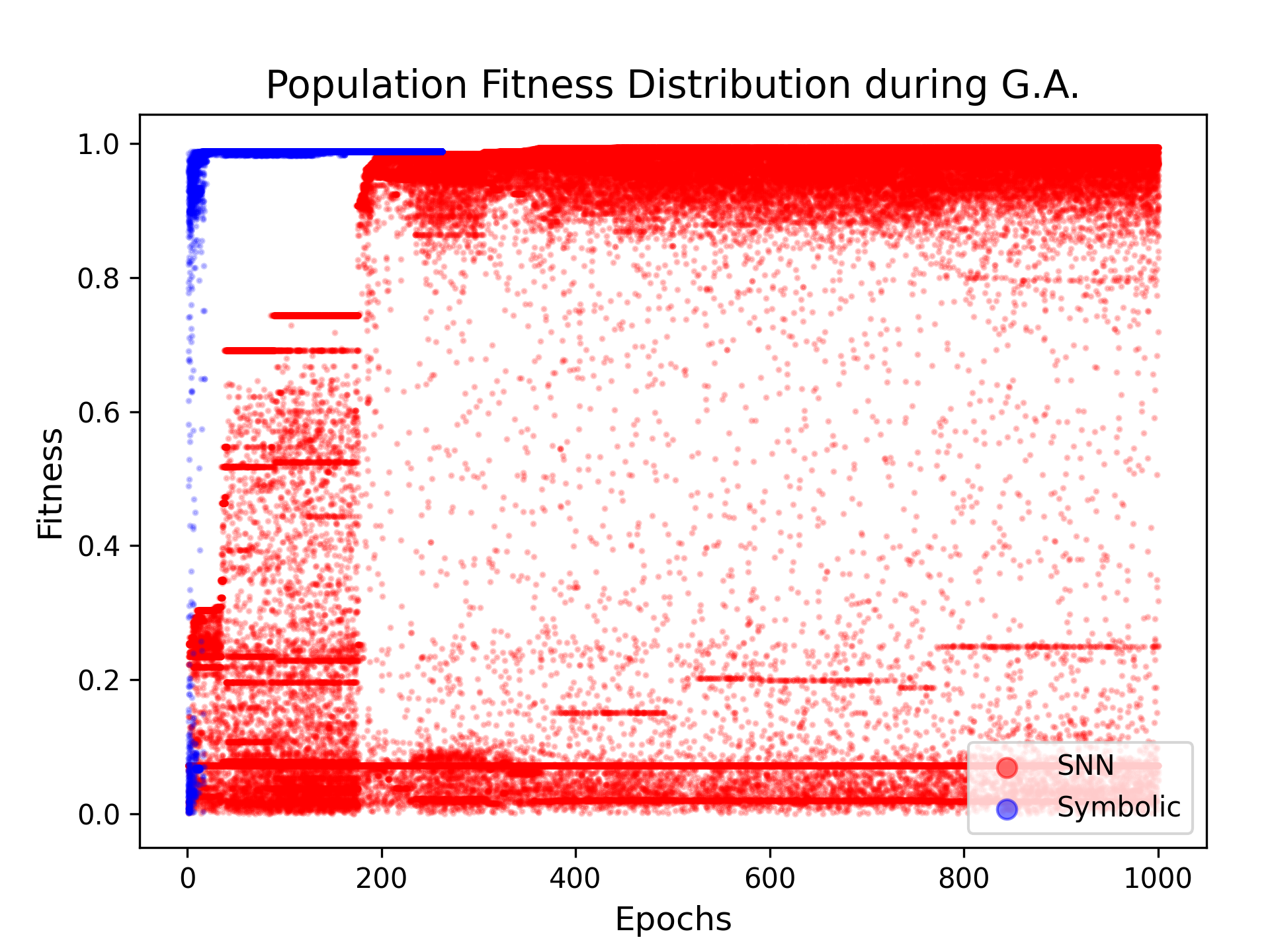}
    \includegraphics[width=.9\linewidth]{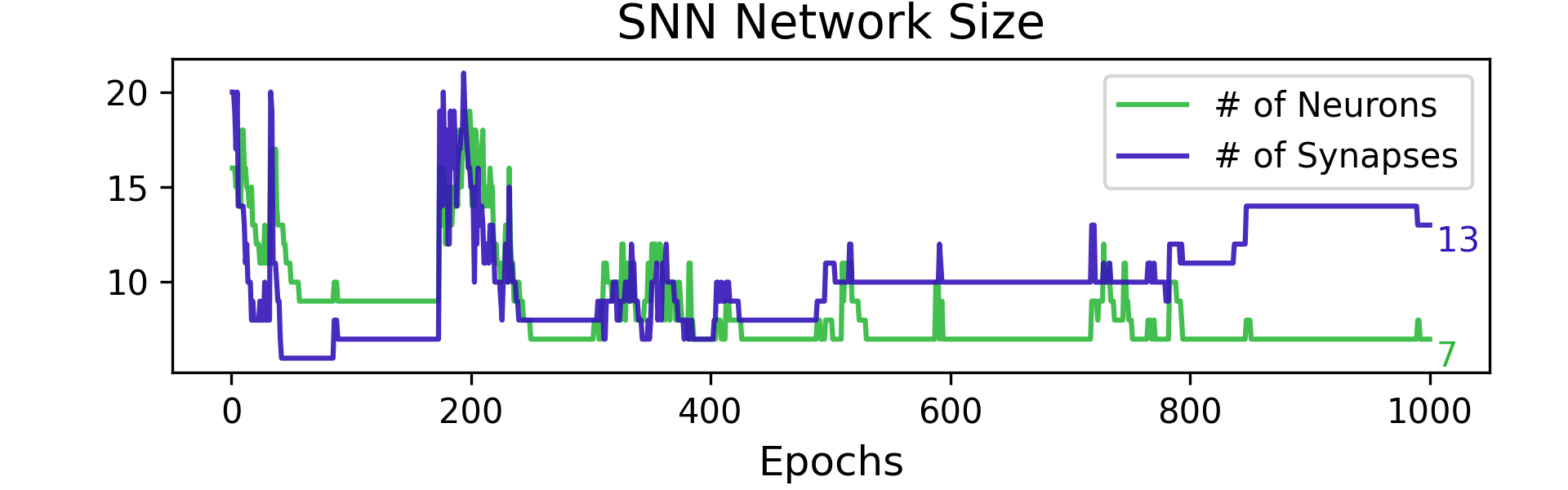}
    \caption{
    \textbf{Top:~}Distribution of fitness scores in the population after each epoch. Each dot represents the fitness of a single simulation in RSSim and is transparent (opacity = 0.2). For EONS, only data from the best run is shown.\\
    \textbf{Bottom:~}Number of neurons and synapses in the best SNN for each epoch of the best EONS run. Note:~Given our encoding/decoding scheme, the minimum number of neurons is 6.
    }
    \label{fig:fit_distribution_netsize}
\end{figure}

\begin{figure}%
    \centering
    \includegraphics[width=.9\linewidth]{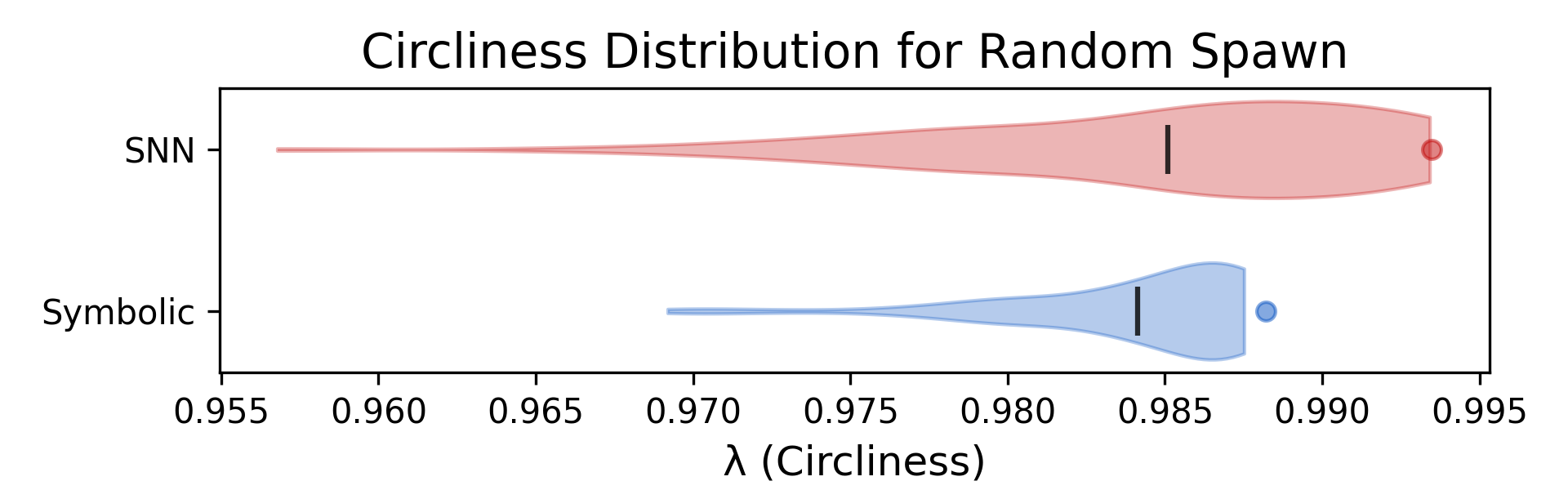}
    \caption{
    Violin plot of circliness in RSSim across 100 seeds for spawn locations (see \eqref{eq:starting_region}). The vertical bar denotes the mean circliness and the dots denote the best overall circliness achieved during training.
    }
    \label{fig:fit_distribution_i0spawn}
\end{figure}

In Figure \ref{fig:bests}, we can see that the distribution of circliness converges within the first few epochs for the symbolic controller optimized with CMA-ES: what is likely the best score possible for $\lambda$ is achieved in the 15th epoch. The CMA-ES run was terminated after 263 epochs because the condition number of the covariance matrix had exceeded the termination threshold: $10^{14}$ in this case. For the SNN controller, we see the variability in the training process represented by the silhouette of minimum and maximum fitnesses. One run surpassed the best overall symbolic circliness score within 100 epochs, but this run did not achieve the highest final $\lambda$ of the five training runs. For 4 out of the 5 runs of EONS, the best SNN achieved a marginally higher score than the best symbolic controller found with CMA-ES.

To evaluate and evolve 100 generations, EONS took 144 minutes to evaluate the SNNs, and CMA-ES took 2 minutes longer.

\begin{figure}
    \centering
    \includegraphics[max height=8em]{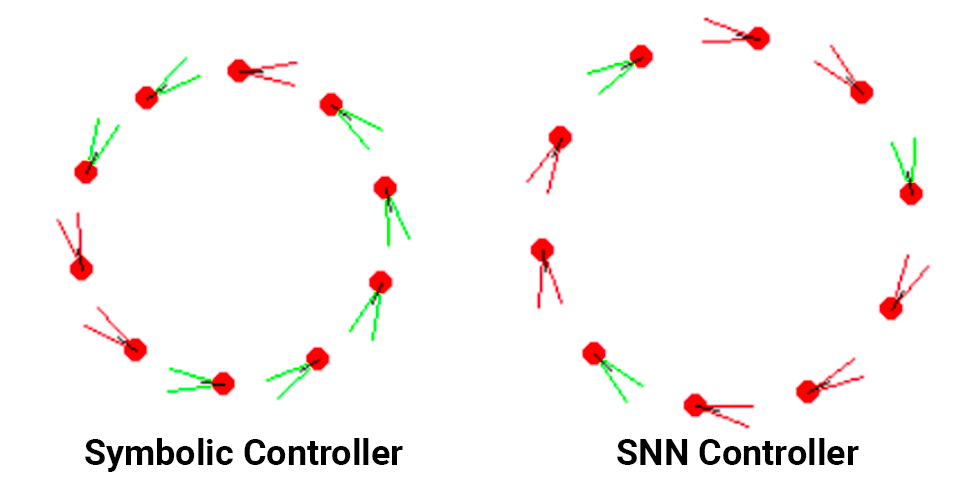}
    \caption{Visualization of milling in RSSim. Green whiskers indicate that the sensor is activated.\\
    \textbf{Left:}~Symbolic Controller Mill.\\
    \textbf{Right:}~SNN Controller Mill.
    }
    \label{fig:sbs}
\end{figure}
\begin{figure}
    \centering
    \includegraphics[width=.9\linewidth]{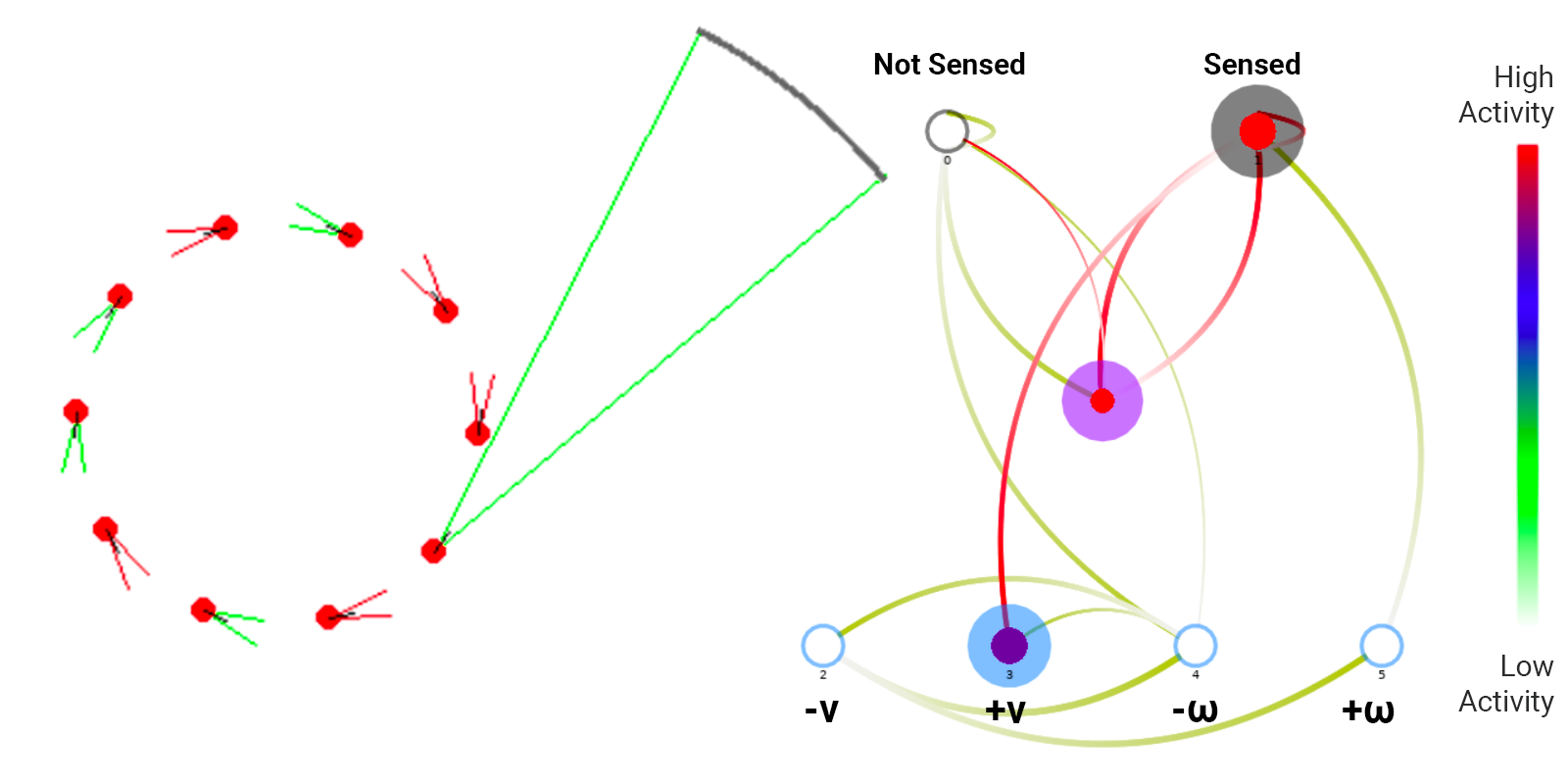}
    \caption{Snapshot of live visualization showing simulation with a selected robot's vision cone (green) shown. The selected robot's live network charge states are shown on the right. The neurons at the top of the graph are the input neurons, and the neurons at the bottom are the outputs, labeled with their contribution to that agent's speed $v_i$ and turning rate $\omega_i$.
    }
    \label{fig:snnviz}
\end{figure}

\subsection{Behavior Analysis}

We examined the behavior of the evolved controllers by observing the simulation, the live network visualization, and the control outputs. In Figure \ref{fig:sbs}, we see that the symbolic controller \eqref{eq:binary_s2a_controller} forms a large, slow-moving, evenly spaced circle. An even number of robots have their sensors activated, and their policy is explicitly known and easy to describe: Turn slightly left (counter-clockwise, out of the circle) if another object comes into view, otherwise turn right (clockwise, into the circle). The difference in turning rates for $h_i = 0$ and $h_i = 1$ is minimal, so as to optimize tangentness, resulting in constant flip-flopping of the sensor.

The behavior for the best evolved SNN controller appears to work in similar fashion, except the mill is moving counter-clockwise. As indicated by the green whiskers, every robot has its sensor activated. The agents follow a similar rule to the symbolic controller, but the speed of each agent fluctuates slightly, even if the sensor state does not change. While there may be more properties of the learned SNN controller, a full analysis is outside the scope of this paper.

\subsection{Discussion}

Although the circliness~$\lambda$~\eqref{eq:circliness} converges considerably faster for the symbolic controller than for the SNN controller, the search and simulation process itself did not take significantly longer for the SNN controller for the same number of epochs. This is despite the fact that simulating the SNN requires additional CPU time to simulate the spikes propagating through the network of the simulated spiking processors. This is evidence that, for small network sizes, the performance overhead of simulating an SNN is negligible, or at least comparable to the time added by the CMA-ES statistical operations.

Figure~\ref{fig:fit_distribution_netsize} shows the network size changes as a result of EONS. We can see that, initially, the network size is set to the values in Table~\ref{ta:eons_params}, but the number of neurons and synapses falls drastically within the first 100 epochs. However, around the 170th epoch, there is a large jump in best network size, accompanied by a gradual shift in fitness distributions. This best net is similar in size to the initial network size set in Table \ref{ta:eons_params}, suggesting that the distribution shift and performance uptick was the result of a randomly initialized network (see Section~\ref{me:eons},~\texttt{random\_factor}). The network size falls again over the next hundred epochs, likely thanks to the multi-objective optimization described in Section~\ref{sn:eons-methods}. Conversely, the growth of the network past the 300th epoch suggests that the additional synapses are responsible for improved circliness scores.

In Figure~\ref{fig:fit_distribution_i0spawn}, we can see that there is some performance degradation when the initial positions are changed, suggesting some slight over-fitting to the fixed initialization position. However, both the SNN and Symbolic controller perform well, with the SNN performing better overall, but with a wider spread compared to that of the Symbolic controller.

At one point, a parameter was incorrectly set such that the SNN's output decoding was binned into three possible outputs per $o_i$, i.e. each agent could only move at ${-0.2, 0, 0.2}$ m/s and turn with ${-2, 0, 2}$ rad/s. Despite this, it was still able to mill with a circliness score of over 0.86, which still results in a milling formation. This may point to SNNs being able to use temporal strategies to compensate for a lack of actuation resolution, but further research is needed.

\section{Conclusions}
In this paper, we investigated use of spiking neural networks (SNNs) for use in a swarm of binary sensing agents. By evolving an SNN to maximize a ``circliness" metric, it is possible to find local interaction rules that lead to a desired group behavior, milling in this case, using SNNs without explicitly defining the structure of the controller.

While this work is limited to simulation, the learned SNN is small enough to easily be deployed on robots, allowing us to test the sim-to-real transfer \cite{vega2023simulate} of the learned behavior in the future.
We hope that this work inspires research on how SNNs can be used to form more complex swarming behaviors.
In these experiments, a candidate controller structure (e.g., the symbolic binary controller) is already known. We would of course like to apply this methodology to more complex problems.
However, to make further progress with complex problems, the training process must become more efficient. 
Possible approaches to this may include back-propagation, but this presents additional challenges such as how to calculate local rewards from the global reward. 
We suggest the milling problem should be used as a benchmark as we work to improve the learning speed, as it requires emergence but is simple to work with and intuitive to understand.

\bibliographystyle{ieeetr}
\bibliography{refs}

\end{document}